\documentclass[10pt,twocolumn,letterpaper]{article}

\usepackage{cvpr}              

\usepackage[dvipsnames]{xcolor}

\definecolor{cvprblue}{rgb}{0.21,0.49,0.74}
\usepackage[pagebackref,breaklinks,colorlinks,citecolor=cvprblue]{hyperref}
\usepackage{multirow}
\usepackage{comment}

\usepackage{colortbl}  
\usepackage{xcolor}

\def\paperID{2051} 
\def\confName{CVPR}
\def\confYear{2024}

\title{Defense without Forgetting: Continual Adversarial Defense with \\ Anisotropic \& Isotropic Pseudo Replay}

\author{Yuhang Zhou$^{1}$, Zhongyun Hua$^{1,2}$\thanks{Zhongyun Hua is the corresponding author.}\\
$^1$Harbin Institute of Technology, Shenzhen,\\
$^2$Guangdong Provincial Key Laboratory of Novel Security Intelligence Technologies\\
    {\tt\small \{23B95105@stu, huazhongyun@\}.hit.edu.cn}
}

\begin{document}
\maketitle
\begin{abstract}
Deep neural networks have demonstrated susceptibility to adversarial attacks. Adversarial defense techniques often focus on one-shot setting to maintain robustness against attack. However, new attacks can emerge in sequences in real-world deployment scenarios. As a result, it is crucial for a defense model to constantly adapt to new attacks, but the adaptation process can lead to catastrophic forgetting of previously defended against attacks. In this paper, we discuss for the first time the concept of continual adversarial defense under a sequence of attacks, and propose a lifelong defense baseline called Anisotropic \& Isotropic Replay (AIR), which offers three advantages: (1) Isotropic replay ensures model consistency in the neighborhood distribution of new data, indirectly aligning the output preference between old and new tasks. (2) Anisotropic replay enables the model to learn a compromise data manifold with fresh mixed semantics for further replay constraints and potential future attacks. (3) A straightforward regularizer mitigates the 'plasticity-stability' trade-off by aligning model output between new and old tasks. Experiment results demonstrate that AIR can approximate or even exceed the empirical performance upper bounds achieved by Joint Training.

\end{abstract}

\section{Introduction}
\label{sec:intro}

\begin{figure}
\centering
\includegraphics[width = 7.5cm]{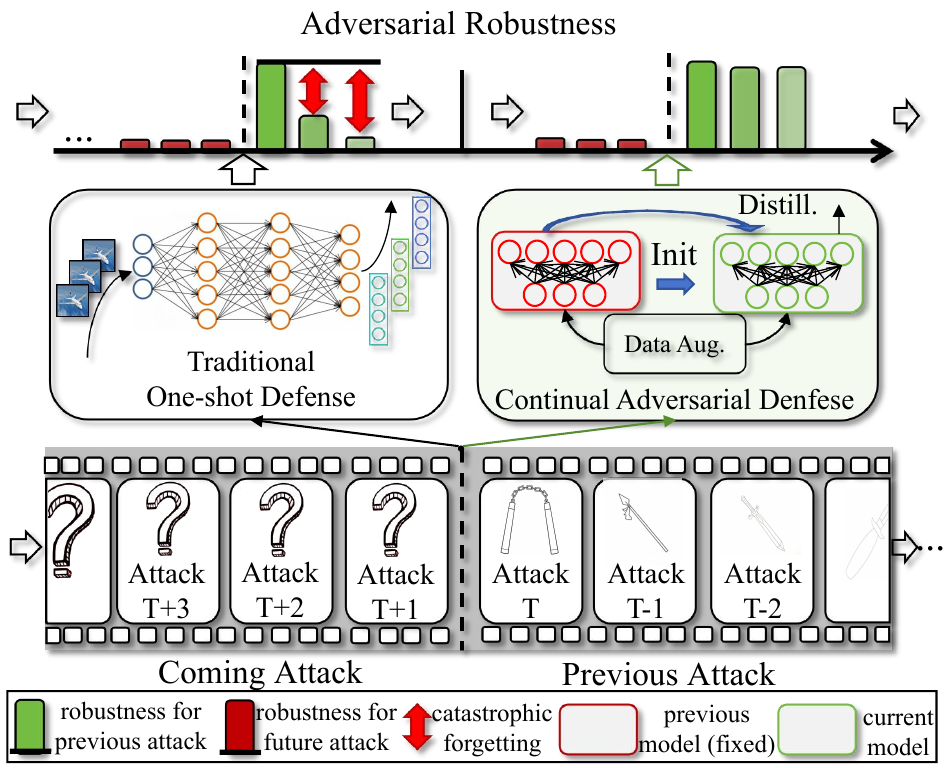}
\caption{The difference between one-shot defense and continual defense. The model diagram on the left presents the one-shot defense studies an isolated Min-Max process and implicitly assumes the potential attack is static. For a continual attack sequence, the indispensable adaptation process introduces additional challenge of catastrophic forgetting of previous attacks. Therefore, a deployable adversarial defense should be a life-one learning task rather than a one-shot task. We propose a self-distillation pseudo-replay baseline to alleviate the catastrophic forgetting against attack sequence, indicated by the model diagram on the right.
}
\label{inv}
\end{figure}

\begin{figure*}
\centering
\includegraphics[width = 14.6cm]{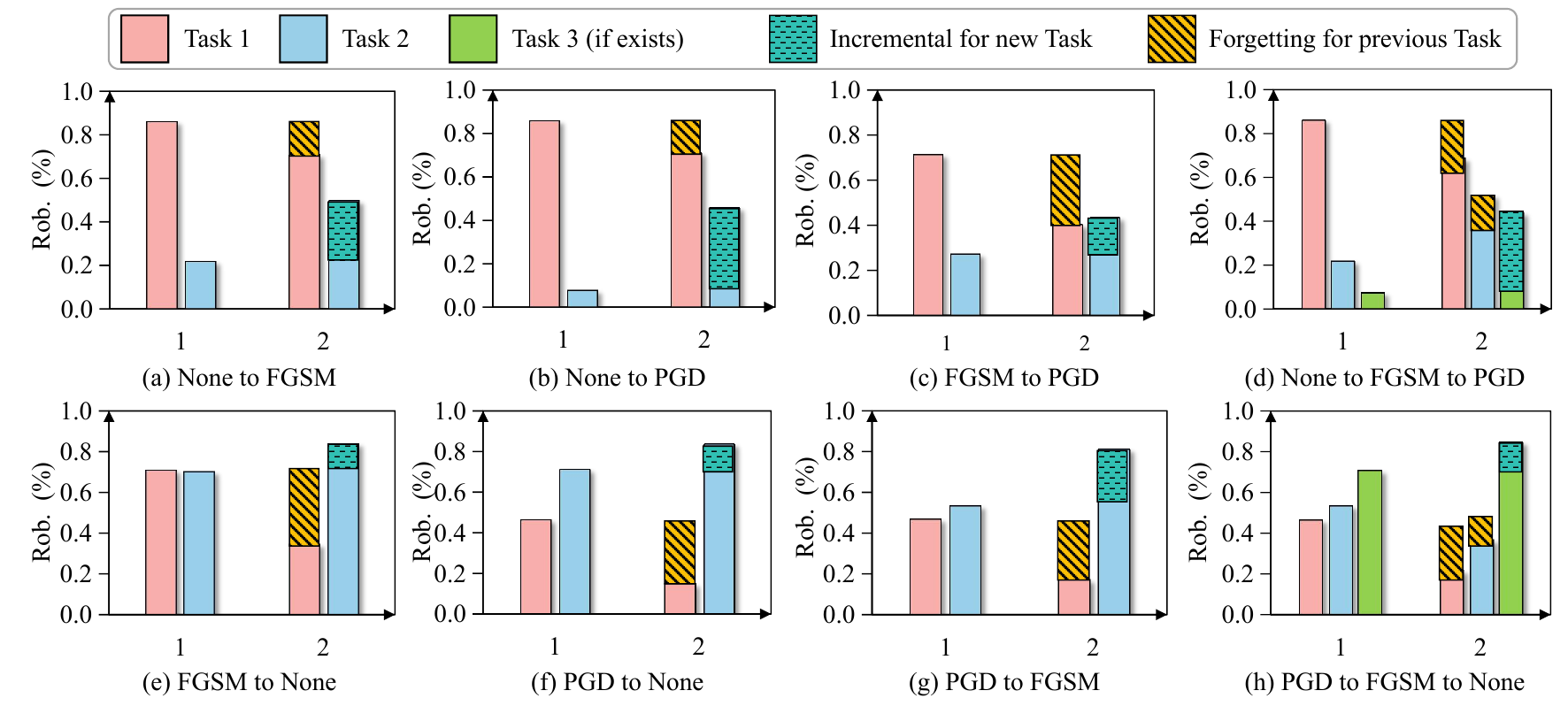}
\caption{Catastrophic forgetting verification of one-shot defense model in continual defense scenario. The horizontal axis can be considered as a timestamp, where time '1' represents the model adapting to \textit{TASK 1}, and time '2' represents the sequential adaptation to all attack tasks in the sequence. \textit{TASK 1} and \textit{TASK 2} depend on the specific sequence. For example, for the sub-first figure, \textit{TASK 1} and \textit{TASK 2} refer to None Attack and FGSM Attack, respectively.}
\label{fig:veri}
\end{figure*}


Plenty of studies have demonstrated that high-performance deep neural networks (DNNs) are vulnerable to adversarial attacks~\cite{szegedy2013intriguing,madry2017towards,moosavi2016deepfool}, indicating that the addition of carefully designed but human-imperceptible perturbations to the inputs of a DNN can easily deceive the network and lead to incorrect predictions, a phenomenon known as adversarial attack. The existence of such attacks presents a significant threat to the deployment of the DNN-based systems. For instance, the elaborate physical adversarial attacks~\cite{kurakin2018adversarial,athalye2018synthesizing,brown2017adversarial} may cause the DNNs-based auto-drive systems to make incorrect judgments, potentially resulting in traffic accidents.
 
To ensure the reliability of DNNs in various scenarios, many defense methods have been proposed to maintain the robustness of DNN against adversarial attacks~\cite{goodfellow2014explaining,liao2018defense,papernot2016distillation,xie2017mitigating,zheng2016improving}. Existing defenses are always limited to the one-shot assumption, which indicates that the model enters a static process after a single defense training stage. However, in real deployment scenarios, new attacks occurs continuously~\cite{goodfellow2014explaining,kurakin2016adversarial,dong2018boosting,madry2017towards,croce2020reliable} and even become a task sequence. As a result, defense model needs to constantly adapt to new attacks for adaptive robustness and turns into a life-long learning~\cite{de2021continual, masana2022class} rather than a one-shot task. Considering that DNNs are easily suffer from catastrophic forgetting~\cite{mccloskey1989catastrophic, de2021continual, masana2022class}, adapting to new attacks is unavoidable to result in suboptimal forgetting of previous one, which poses a new threat to the reliability of DNNs.  Figure~\ref{inv} illustrates the difference between one-shot and continual
defense. 

To demonstrate  this challenge, we first explore the continual challenge of DNNs' adversarial robustness under continual attack sequences. We validate the catastrophic forgetting of standard adversarial training~\cite{madry2017towards} on the attack sequences consisting of two and three attacks, as shown in Figure~\ref{fig:veri}. The experiment is conducted on the CIFAR10 dataset with the backbone being Wide-ResNet-34~\cite{zagoruyko2016wide}. The validation experiment includes two settings: 'from easy to difficult' (i.e., FGSM to PGD) and 'from difficult to easy' (i.e., PGD to FGSM). For more validation experiments under different attack principles, varying attack intensities, and transferring between black and white box attacks), please refer to Section~\textcolor{red}{1} of the supplementary materials. Obviously, adversarial training suffers from significant catastrophic forgetting under all attack sequences, and the forgetting becomes more severe under the 'difficult to easy' attack sequence. These results confirm our concern about DNNs' forgetting robustness to previous attacks as they constantly adapt to new ones, highlighting the need for defense with continual adversarial robustness. 


To achieve robustness against continual adversarial, a potential solution is to employ Joint Training with all sequential attacks or fine-tune a pre-trained robust model with new attacks. However, the Joint Training strategy have challenges as the previous attack data of pre-trained models may not access due to privacy protection and legal restrictions, making Joint Training difficult. Additionally, since new attacks are constantly emerging, the training cost linearly increases as the increase of attack sequence, making it inefficient to train all the data every time. For the fine-tuning strategy, poor plasticity can lead to forgetting or insufficient learning. Therefore, defense in the source-free continual paradigm~\cite{de2021continual, masana2022class} is necessary under the scenario of data privacy, adversarial robustness and life-long learning.


In this paper, we propose \textit{Anisotropic \& Isotropic  Replay (AIR)} as a baseline for continual adversarial defense. AIR combines isotropy and anisotropy data augmentation to alleviate catastrophic forgetting in continual adversarial defense scenarios within a self-distillation pseudo replay paradigm. The isotropic stochastic augmentation is beneficial for breaking specific adversarial patterns, and an isolated learning objectives is obtained for pseudo-replay data to prevent pattern collapse and self-contradiction, implicitly constraining the consistency between the new model and the previous model in the neighbor distribution of new attacks. This alignment  indirectly aligns the model's output of new and previous attacks, considering that all adversarial data can be regarded as neighborhood samples of the raw data. The anisotropic mix-up augmentation provides the model with richer fusion semantics and connects the manifolds of previous and new attacks, originally spaced more apart. To further optimize the trade-off between plasticity and stability, we introduce an intuitive regularizer to optimize the model's generalization by constraining the hidden-layer feature of new attacks and pseudo-replay attacks to be mapped to the same feature cluster. 

Our contribution can be summarized as:

$\bullet$ We first discuss, validate, and analyze the catastrophic forgetting challenge of adversarial robustness under continual attack sequences threat.

$\bullet$ We tackle the twin challenge of adversarial robustness and continual learnability by proposing AIR as an efficient self-training reply baseline. Through the pseudo self-distillation, parameters with similar activation for new and previous attacks are found and retained. AIR can also achieve an implicit chain consistency.

$\bullet$ We evaluate the performance of several classic and plug-and-play continual learning methods for continual adversarial attacks by combining them with adversarial training. Qualitative and quantitative experiments verify the feasibility and superiority of our AIR.

\section{Related Work}
\label{sec:rel}
\subsection{Adversarial Attack \& Defense}
\textbf{Adversarial Attack.} Attacks on DNN models can be catrgorized into white box and black box attacks based on whether the attacker can access the target model or not. In the white box attack, the adversary can fully query and access various aspects of the target model, such as the model parameters, structure, and gradients. Mainstream white box attacks include gradient-based techniques (e.g., FGSM~\cite{goodfellow2014explaining}, BIM~\cite{kurakin2016adversarial}, MIM~\cite{dong2018boosting}, PGD~\cite{madry2017towards}, and AA~\cite{croce2020reliable}), conditional optimization-based techniques (e.g., CW~\cite{carlini2017towards2}, OnePixel~\cite{su2019one}), and classifier perturbations-based techniques (e.g., DeepFool~\cite{moosavi2016deepfool}). On the other hand, black box attack occurs when the attacker has limited knowledge about the target model. This category can be further divided into score-based, decision-based, and transfer-based attacks. In score-based black box attack, the attacker can access to the probabilities (e.g., zoo~\cite{chen2017zoo}). Decision-based black box attacks operate under the constraint that the attacker can only obtain the one-hot prediction (e.g., Boundary attacks~\cite{brendel2017decision}). Transfer-based attack~\cite{dong2018boosting,dong2019evading,lin2019nesterov} typically involve crafting adversarial attacks using a substitute model, commonly employed to evaluate the adversarial robustness of DNNs.


\noindent\textbf{Adversarial Defense.} To maintain adversarial robustness under attacks, early adversarial defense are often heuristic, including input transformation~\cite{xie2017mitigating}, model ensemble~\cite{bagnall2017training}, adversarial denoiser~\cite{shen2017ape}. However, most of these models have been proven to benefit from unreliable obfuscated gradients~\cite{athalye2018obfuscated}. Recently, adversarial training (AT)~\cite{goodfellow2014explaining,madry2017towards,jia2022adversarial} and defensive distillation~\cite{goldblum2020adversarially,wang2019improving,wang2019improving,zhu2021reliable} have become mainstream defense due to their essential robustness, and the latest research primarily focus on exploring their potential. In adversarial training, Jia~$et~al.$~\cite{jia2022adversarial} proposed a learnable strategy, Mustafa~$et~al.$~\cite{mustafa2020deeply} enhanced adversarial robustness by perturbing feature representations, and Pang~$et~al.$~\cite{pang2020bag} fully utilized tricks to maximize the potential of AT. For defensive distillation, Wang $et~al.$~\cite{wang2021agkd} introduced a bidirectional metric learning framework guided by an attention mechanism, and Zi~$et~al.$~\cite{zi2021revisiting} proposed to fully exploit soft labels generated by a robust teacher model. The aforementioned defenses are limited to the one-shot settings and cannot adapt to new attack, resulting in insufficient robustness for the potential attack sequences. Some latest attempts to address this adaptive challenge include Test Time Adaptation Defense (TTAD)~\cite{shi2021online,yang2022adaptive}, which considers  continual adaptation to new attacks. However, TTAD only focuses on adaptation to unlabeled attacks on test data based on the current model, and ignores to alleviate the catastrophic forgetting of previous attacks. In this study, we propose a novel continual defense task where the defender considers both new and previous attacks.



\subsection{Continual \& Incremental Learning}
DNNs tend to suffer from catastrophic forgetting, where adaptation to new tasks leads to a drop of performance for previous tasks. Continual learning~\cite{masana2022class}, also known as incremental learning~\cite{de2021continual} and life-long learning~\cite{parisi2019continual}, is considered a potential solution to atastrophic forgetting. In continual learning, tasks are roughly divided into 'Incremental task learning', 'Incremental class learning', and 'Incremental domain learning' based on the differences between new and previous tasks. For continual learning models, methods can be roughly categorized into replay-based~\cite{rolnick2019experience,chaudhry2019continual}, parameter-isolation-based~\cite{aljundi2017expert,xu2018reinforced}, and regularization-based~\cite{li2017learning,kirkpatrick2017overcoming,rannen2017encoder} methods, depending on how the task-specific information is stored and used throughout the sequential learning process. Replay-based methods can be further be divided into explicit replay~\cite{chaudhry2019continual} and pseudo replay~\cite{robins1995catastrophic}. Regularization-based methods can further be divided into data-focused~\cite{li2017learning} and prior-focused methods~\cite{kirkpatrick2017overcoming}. Parameter-isolation-based methods can be further divided into module allocation~\cite{aljundi2017expert}) and additional modules~\cite{kim2023achieving}. Despite several attempts on basic tasks, continuous learning scenarios have not yet been deeply explored in adversarial defense. With the increasing variety of attacks, reliable DNN models necessitate  adaptable life-long defense against attack sequences.



\begin{figure*}
\centering
\vspace{-0.1cm}
\includegraphics[width = 14cm]{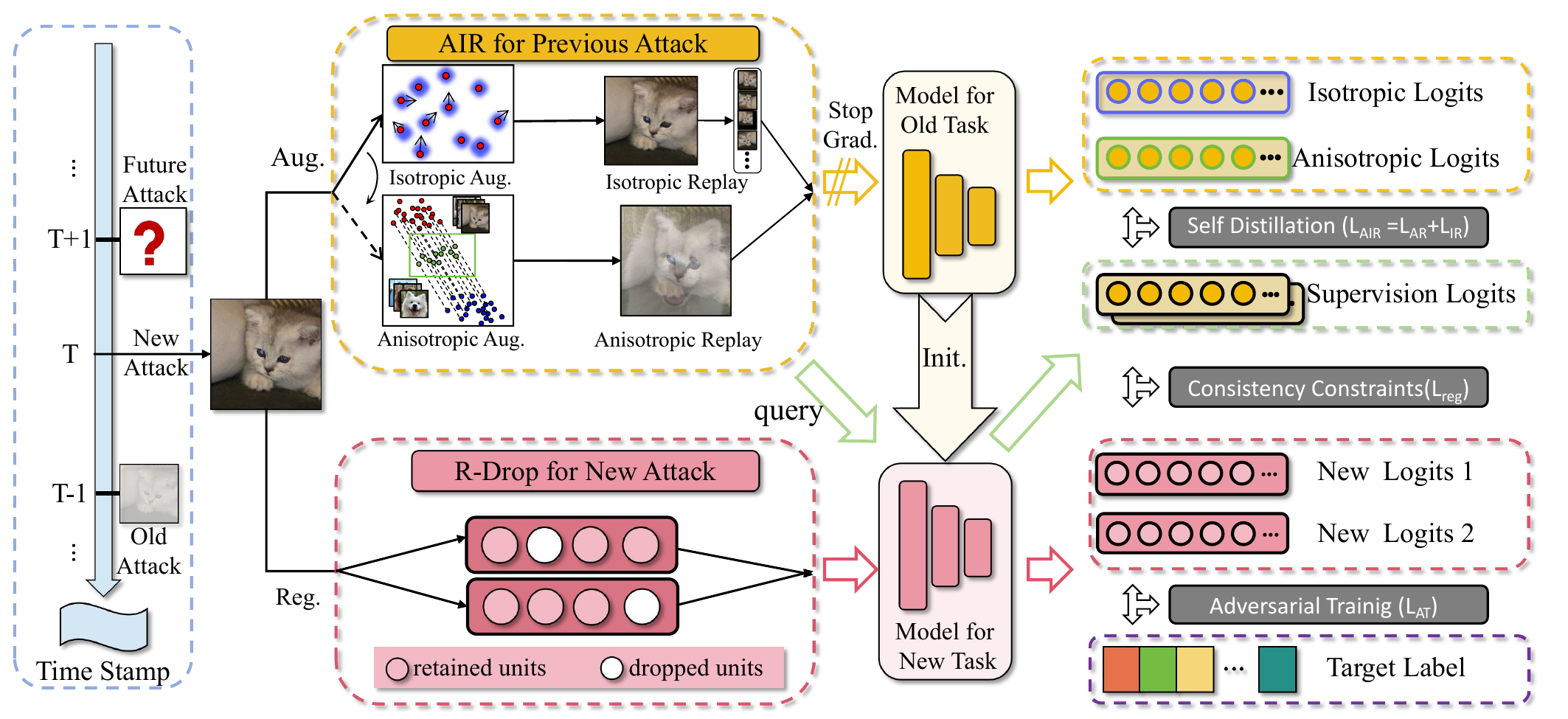}
\caption{Framework of our AIR. The upper module (in yellow block) consists of the anisotropic replay module and isotropic replay module, aiming to maintain the memory of old tasks. The lower module (in red block) is the vanilla adversarial training with R-Drop for new attacks. The three main loss functions are highlighted in the gray circular box.}
\label{fig:air}
\end{figure*}


\section{Methodology}\label{sec3}
\subsection{Overview}
Continual adversarial defense resembles 'incremental domain learning' more than 'incremental task learning' or 'incremental class learning'~\cite{masana2022class}. On one hand, the output space of the attack sequence is fixed for one dataset, allowing the defense model for each attack to share a common classifier. On the other hand, even though the distribution at the feature level may shift, the low-level semantics of adversarial samples remain invariant, which means that the adversarial samples of cats are always visually the same as cats (constrained by the definition of adversarial samples) and can be considered as the neighbor of the raw data. As a result, complex prior knowledge and large memory may not be necessary to obtain replay data. 

However, directly utilizing new attack data as pseudo replay data may be insufficient. On the one hand, new data needs to be mapped to real labels. Simultaneously aligning the output of the new data with the output of the teacher model and the real label may lead the model to be trapped in a self-contradiction dilemma. On the other hand, the new data are still adversarial, making it challenging to for the defense model to fit them without forgetting. Hence, we propose a composite data augmentation scheme to establish an efficient self-distillation pseudo replay paradigm.


\subsection{Task Definition.} \label{def} 
We first provide a definition of continual adversarial defense. In continual adversarial defense, the model learns an attack sequence $\mathcal{A}=\{A_1, ...A_t, ... A_N\}$ one by one, and the attack ID is available during both training and testing stages. We assume that each attack $A_t$ has a manually labeled training set and test set, denoted as $A_t^{train}=\{(x_t^i, y_t^i), i=1,..., n_t^{train}\}$, where $y_t^i$ is the real label, and $n_t^{train}$ is the number of the training data. The test set can be defined as $A_t^{test}$ in the similar way. A reliable defense model should learn a new attack $A_t$ without forgetting the previous attacks $\{A_1, ...A_{t-1}\}$.

\subsection{Self-distillation Pseudo Replay} \label{sdpr}
By aligning the outputs of the current and previous models on pseudo replay data, the current model can learn the mapping preferences of the previous model, indirectly maintaining the mapping relationship between the previous data-label pair. Generally, assuming we have a model $f_{w_{t}}$ parameterized by $w_{t}$ at time $t$, the self distillation pseudo replay can be formalized as:
\begin{equation}
\begin{aligned}
\mathcal{L}=\mathcal{L}_{vanilla} +\mathcal{L}_{dis},
\end{aligned}
\label{eq2}
\end{equation}
where $\mathcal{L}_{vanilla}$ is the classification loss of the current model for new data, commonly known as the cross entropy loss. In continual adversarial defense, $\mathcal{L}_{vanilla}$ refers to the vanilla adversarial training loss for a new attack~\cite{madry2017towards} and we represent it as $\mathcal{L}_{AT}$ in the following text.


The $\mathcal{L}_{dis}$ represents the self-distillation loss from the previous model to the current model on pseudo replay data, and can be formalized as:
\begin{equation}
\begin{aligned}
\mathcal{L}_{dis} = D(f_{w_{t-1}}(X_t'), f_{w_{t}}(X_t')),
\end{aligned}
\label{eq2}
\end{equation}
where $D$ is the diversity measure, commonly known as the KL Divergence, $f_{w_{t-1}}$ is the optimal previous model at time $t-1$, and $X_t'$ is the pseudo data at time $t$.

\subsection{Isotropic Pseudo Replay} \label{ir}
To align the outputs of the new and old models, we create independent data for pseudo replay based on the current adversarial samples. Assume that we already have a certain batch of data $X_t$, the neighborhood samples of new data can be obtained by:
\begin{equation}
\begin{aligned}
X_t^{IR} = \mathcal{T}(X_t + \lambda \cdot r),
\end{aligned}
\label{eq2}
\end{equation}
where $\lambda$ is a hyper-parameter, $r$ is a stochastic perturbation sampled from the Gaussian distribution, and $ \mathcal{T}$ is a random augmentation operator that includes random rotation, cropping, flipping, and erasing. In this 'perturbation on attack' way, specific distributions (e.g., pixel or texture \cite{maiya2021frequency,huang2022adversarially}) in the adversarial perturbation are somehow broken, wakening the attack ability. Simultaneously, independent pseudo replay data unrealted to the current attack is obtained. The augmented pseudo replay data does not deviate from the raw semantics, and its potential label is also fixed. Hence it is called isotropic replay (IR), and the IR loss can be formalized as:
\begin{equation}
\begin{aligned}
\mathcal{L}_{IR} = KL\_Div(f_{w_t}(X_t^{IR}), f_{w_{t-1}}(X_t^{IR})).
\end{aligned}
\label{eq2}
\end{equation}

\subsection{Anisotropic Pseudo Replay} \label{ar}

To further obtain a compact and uniform manifold of the replay data and ensure a nontrivial solution, we introduce an anisotropic data augmentation scheme: $mix-distill$, which evolves from $mixup$ \cite{zhang2017mixup}. We randomly shuffle the current batch data $X_t$ to obtain a new batch of data $X_t^{shuffle}$:
\begin{equation}
\begin{aligned}
X_t^{AR} = \alpha \cdot X_t + (1-\alpha) \cdot x_t^{shuffle},
\end{aligned}
\label{eq2}
\end{equation}
where $\alpha$ is a stochastic mixing weight sampled from $U[0.3, 0.7]$. The label for the replay is also required. The pseudo label for the mixed data is obtained in the same way as in the vanilla \textit{mixup}. However, real labels are agnostic in pseudo replay framework and we attempt to obtain the mixed labels in two ways:

\noindent\textbf{Mixed query labels.} Referring to the standard $mixup$, we collect the output logits of the teacher model (previous model) for the two components of the mixed data, and mix them with the same weight as the data:
\begin{equation}
\begin{aligned}
y_{dis} = \alpha \cdot f_{w_{t-1}}(X_t) + (1-\alpha) \cdot f_{w_{t-1}}(X_t^{shuffle}).
\end{aligned}
\label{eq2}
\end{equation}

However, the label mixing strategy in the standard $mixup$ sometimes leads to training collapse. This is because pseudo labels are inherently inaccurate, and mixing suboptimal pseudo labels leads to error accumulation, making it difficult to align the preference of teachers and students. The nonlinearity of the model may also amplify the mapping shift of the pseudo-replay self-distillation model.

\noindent\textbf{Query label of the mixed data.} Based on the above analysis, we directly align the output preferences of the teacher-student model for mixed samples, and the AR losses can be expressed as
\begin{equation}
\begin{aligned}
\mathcal{L}_{AR} = KL\_Div(f_{w_{t-1}}(X_{AR}), f_{w_{t}}(X_{AR})).
\end{aligned}
\label{eq2}
\end{equation}

The nature of 'anisotropy' is reflected not only in the feature mixing provided by pixel-level interpolation (such as the mixing of $lion$ and $tiger$, which may introduce new semantics like $liger$), but also in the indirect combination of supervise labels. In this way, the intra-class gap is initially filled, and the data manifold becomes uniform. This also provides the model with the ability to generalize to unfamiliar and unseen semantics, as such semantics may arise in future tasks. More specifically, adversarial samples may have already introduced new semantics to the original data, and the mixing between adversarial samples further explores a richer internal maximization process.

\begin{table*}[]
\footnotesize
\centering
\setlength{\tabcolsep}{1.5mm}{
\begin{tabular}{cccccccccccccc}
\toprule
\multicolumn{14}{c}{Transfer between two attacks}                                                                                                                        \\ \toprule
\multicolumn{1}{c|}{}                           & \multicolumn{1}{c|}{}                        & \multicolumn{2}{c|}{None to FGSM}                                   & \multicolumn{2}{c|}{FGSM to None}                          & \multicolumn{2}{c|}{None to PGD}                           & \multicolumn{2}{c|}{PGD to None}                           & \multicolumn{2}{c|}{FGSM to PGD}             & \multicolumn{2}{c}{PGD to FGSM}             \\
\multicolumn{1}{c|}{\multirow{-2}{*}{Datasets}} & \multicolumn{1}{c|}{\multirow{-2}{*}{Tasks}} & Task 1                        & \multicolumn{1}{c|}{Task 2}         & Task 1               & \multicolumn{1}{c|}{Task 2}         & Task 1               & \multicolumn{1}{c|}{Task 2}         & Task 1               & \multicolumn{1}{c|}{Task 2}         & Task 1         & \multicolumn{1}{c|}{Task 2} & Task 1               & Task 2               \\ \midrule
\multicolumn{1}{c|}{}                           & \multicolumn{1}{c|}{Vanilla AT \cite{madry2017towards}}                 & 95.18                         & \multicolumn{1}{c|}{98.55}          & 83.97                & \multicolumn{1}{c|}{98.86}          & 94.22                & \multicolumn{1}{c|}{90.01}          & 3.72                 & \multicolumn{1}{c|}{98.59}          & 96.48          & \multicolumn{1}{c|}{94.71}  & 2.56                 & 96.96                \\
\multicolumn{1}{c|}{}                           & \multicolumn{1}{c|}{EWC \cite{kirkpatrick2017overcoming}}                     & 98.83                         & \multicolumn{1}{c|}{96.63}          & 98.18                & \multicolumn{1}{c|}{97.85}          & 97.35                & \multicolumn{1}{c|}{87.32}          & 91.97                & \multicolumn{1}{c|}{98.85}          & 95.26          & \multicolumn{1}{c|}{\textbf{95.86}}  & 94.77                & 96.90                \\
\multicolumn{1}{c|}{}                           & \multicolumn{1}{c|}{Feat. Extraction \cite{li2017learning}}      & 98.16                         & \multicolumn{1}{c|}{89.23}          & 97.46                & \multicolumn{1}{c|}{98.80}          & 12.72                & \multicolumn{1}{c|}{11.35}          & 95.23                & \multicolumn{1}{c|}{98.80}          & 96.94          & \multicolumn{1}{c|}{73.61}  & 95.23                & 97.93                \\
\multicolumn{1}{c|}{}                           & \multicolumn{1}{c|}{LFL \cite{jung2016less}}                     & 98.85                         & \multicolumn{1}{c|}{97.02}          & 90.54                & \multicolumn{1}{c|}{98.80}          & 97.32                & \multicolumn{1}{c|}{87.52}          & 33.84                & \multicolumn{1}{c|}{98.71}          & 95.84          & \multicolumn{1}{c|}{91.87}  & 25.05                & \textbf{98.40}                \\
\multicolumn{1}{c|}{}                           & \multicolumn{1}{c|}{\textbf{AIR (ours)}}            & \textbf{99.37}                & \multicolumn{1}{c|}{\textbf{98.84}} & \textbf{98.18}       & \multicolumn{1}{c|}{\textbf{98.84}} & \textbf{98.89}       & \multicolumn{1}{c|}{\textbf{94.26}} & \textbf{95.93}       & \multicolumn{1}{c|}{\textbf{99.06}} & \textbf{97.45} & \multicolumn{1}{c|}{95.67}  & \textbf{96.25}                & 97.93                \\
\multicolumn{1}{c|}{\multirow{-6}{*}{MNIST}}    & \multicolumn{1}{c|}{\cellcolor{lightgray} Joint Training  \cite{li2017learning}}             & \cellcolor{lightgray}99.11 & \multicolumn{1}{c|}{\cellcolor{lightgray}98.52}          & \cellcolor{lightgray}98.52                & \multicolumn{1}{c|}{\cellcolor{lightgray}99.11}          & \cellcolor{lightgray}99.35                & \multicolumn{1}{c|}{\cellcolor{lightgray}95.44}          & \cellcolor{lightgray}95.44                & \multicolumn{1}{c|}{\cellcolor{lightgray}99.35}          & \cellcolor{lightgray}96.72          & \multicolumn{1}{c|}{\cellcolor{lightgray}94.29}  & \cellcolor{lightgray}94.29                & \cellcolor{lightgray}96.72                \\ \midrule
\multicolumn{1}{c|}{}                           & \multicolumn{1}{c|}{Vanilla AT \cite{madry2017towards}}                 & 70.60                         & \multicolumn{1}{c|}{49.30}          & 34.90                & \multicolumn{1}{c|}{83.83}          & 71.09                & \multicolumn{1}{c|}{45.52}          & 15.19                & \multicolumn{1}{c|}{83.59}          & 34.90          & \multicolumn{1}{c|}{35.21}  & 17.14                & 60.24                \\
\multicolumn{1}{c|}{}                           & \multicolumn{1}{c|}{EWC \cite{kirkpatrick2017overcoming}}                     & 72.66                         & \multicolumn{1}{c|}{49.17}          & \textbf{43.85}                & \multicolumn{1}{c|}{82.62}          & 69.38                & \multicolumn{1}{c|}{41.46}          & 30.25                & \multicolumn{1}{c|}{61.70}          & 48.63          & \multicolumn{1}{c|}{40.53}  & 24.44                & 45.18                \\
\multicolumn{1}{c|}{}                           & \multicolumn{1}{c|}{Feat. Extraction \cite{li2017learning}}      & 67.69                         & \multicolumn{1}{c|}{35.11}          & 45.27                & \multicolumn{1}{c|}{82.13}          & 40.04                & \multicolumn{1}{c|}{30.90}          & \textbf{45.54}                & \multicolumn{1}{c|}{75.02}          & 52.85          & \multicolumn{1}{c|}{24.88}  & 42.51                & 44.54                \\
\multicolumn{1}{c|}{}                           & \multicolumn{1}{c|}{LFL \cite{jung2016less}}                     & 74.23                         & \multicolumn{1}{c|}{50.17}          & 42.77                & \multicolumn{1}{c|}{78.59}          & 67.31                & \multicolumn{1}{c|}{42.76}          & 28.27                & \multicolumn{1}{c|}{\textbf{80.59}}          & 51.98          & \multicolumn{1}{c|}{43.30}  & 24.18                & 46.71                \\
\multicolumn{1}{c|}{}                           & \multicolumn{1}{c|}{\textbf{AIR (ours)}}            & \textbf{76.73}                         & \multicolumn{1}{c|}{\textbf{51.48}}          & 42.32                & \multicolumn{1}{c|}{\textbf{82.85}}          & \textbf{75.53}                & \multicolumn{1}{c|}{\textbf{45.14}}          & 41.21                & \multicolumn{1}{c|}{77.02}          & \textbf{53.39}          & \multicolumn{1}{c|}{\textbf{44.12}}  & \textbf{43.00}                & \textbf{52.26}                \\
\multicolumn{1}{c|}{\multirow{-6}{*}{CIFAR10}}  & \multicolumn{1}{c|}{\cellcolor{lightgray}Joint Training \cite{li2017learning}}             & \cellcolor{lightgray}86.10                         & \multicolumn{1}{c|}{\cellcolor{lightgray}47.65}          & \cellcolor{lightgray}57.65                & \multicolumn{1}{c|}{\cellcolor{lightgray}86.10}          & \cellcolor{lightgray}72.58                & \multicolumn{1}{c|}{\cellcolor{lightgray}44.86}          & \cellcolor{lightgray}44.86                & \multicolumn{1}{c|}{\cellcolor{lightgray}72.58}          & \cellcolor{lightgray}49.81          & \multicolumn{1}{c|}{\cellcolor{lightgray}42.56}  & \cellcolor{lightgray}42.56                & \cellcolor{lightgray}49.81                \\ \midrule
\multicolumn{1}{c|}{}                           & \multicolumn{1}{c|}{Vanilla}                 & 42.27                             & \multicolumn{1}{c|}{20.67}              & 25.98                    & \multicolumn{1}{c|}{50.26}              & 40.58                    & \multicolumn{1}{c|}{17.31}              & 20.21                    & \multicolumn{1}{c|}{47.47}              & 24.08              & \multicolumn{1}{c|}{19.03}      & 20.89                    & 30.47                    \\
\multicolumn{1}{c|}{}                           & \multicolumn{1}{c|}{EWC \cite{kirkpatrick2017overcoming}}                     & 50.04                             & \multicolumn{1}{c|}{22.43}              & \textbf{29.13}                    & \multicolumn{1}{c|}{45.12}              & \textbf{48.45}                    & \multicolumn{1}{c|}{16.61}              & 19.21                    & \multicolumn{1}{c|}{44.66}              & 22.98              & \multicolumn{1}{c|}{18.00}      & 20.16                    & 24.32                    \\
\multicolumn{1}{c|}{}                           & \multicolumn{1}{c|}{Feat. Extraction \cite{li2017learning}}      & 37.02                             & \multicolumn{1}{c|}{8.35}              & 23.62                    & \multicolumn{1}{c|}{47.68}              & 11.46                    & \multicolumn{1}{c|}{4.96}              & 20.70                    & \multicolumn{1}{c|}{41.42}              & 23.63              & \multicolumn{1}{c|}{18.22}      & 19.54                    & 24.08                    \\
\multicolumn{1}{c|}{}                           & \multicolumn{1}{c|}{LFL \cite{jung2016less}}                     & 28.61                             & \multicolumn{1}{c|}{15.30}              & 37.48                    & \multicolumn{1}{c|}{49.06}              & 19.19                    & \multicolumn{1}{c|}{13.36}              & 20.08                    & \multicolumn{1}{c|}{43.62}              & 25.49              & \multicolumn{1}{c|}{15.77}      & 19.19                    & 23.85                    \\
\multicolumn{1}{c|}{}                           & \multicolumn{1}{c|}{\textbf{AIR (ours)}}            & \textbf{50.77}                             & \multicolumn{1}{c|}{\textbf{24.32}}              & 27.47                    & \multicolumn{1}{c|}{\textbf{50.67}}              & 47.88                    & \multicolumn{1}{c|}{\textbf{21.41}}              & \textbf{22.05}                    & \multicolumn{1}{c|}{\textbf{45.61}}              & \textbf{27.59}              & \multicolumn{1}{c|}{\textbf{23.19}}      & \textbf{23.40}                    & \textbf{27.51}                    \\
\multicolumn{1}{c|}{\multirow{-6}{*}{CIFAR100}} & \multicolumn{1}{c|}{\cellcolor{lightgray}Joint Training \cite{li2017learning}}             & \cellcolor{lightgray}56.44                             & \multicolumn{1}{c|}{\cellcolor{lightgray}35.88}              & \cellcolor{lightgray}35.88                    & \multicolumn{1}{c|}{\cellcolor{lightgray}56.44}              & \cellcolor{lightgray}46.01                    & \multicolumn{1}{c|}{\cellcolor{lightgray}22.54}              & \cellcolor{lightgray}22.54                    & \multicolumn{1}{c|}{\cellcolor{lightgray}46.01}              & \cellcolor{lightgray}35.27              & \multicolumn{1}{c|}{\cellcolor{lightgray}21.45}      & \cellcolor{lightgray}21.45                    & \cellcolor{lightgray}35.27                    \\ \bottomrule
\end{tabular}}
\caption{Adaptation between two attacks for different defense methods. Each method is combined with vanilla adversarial training and the Joint Training is considered to be the empirical upper bound marked with a gray background. The best performance, excluding Joint Training, is highlighted in bold.}
\label{tab1}
\end{table*}



\subsection{Regularization for the Trade-off} \label{reg}

\begin{figure}
\centering
\includegraphics[width = 7cm]{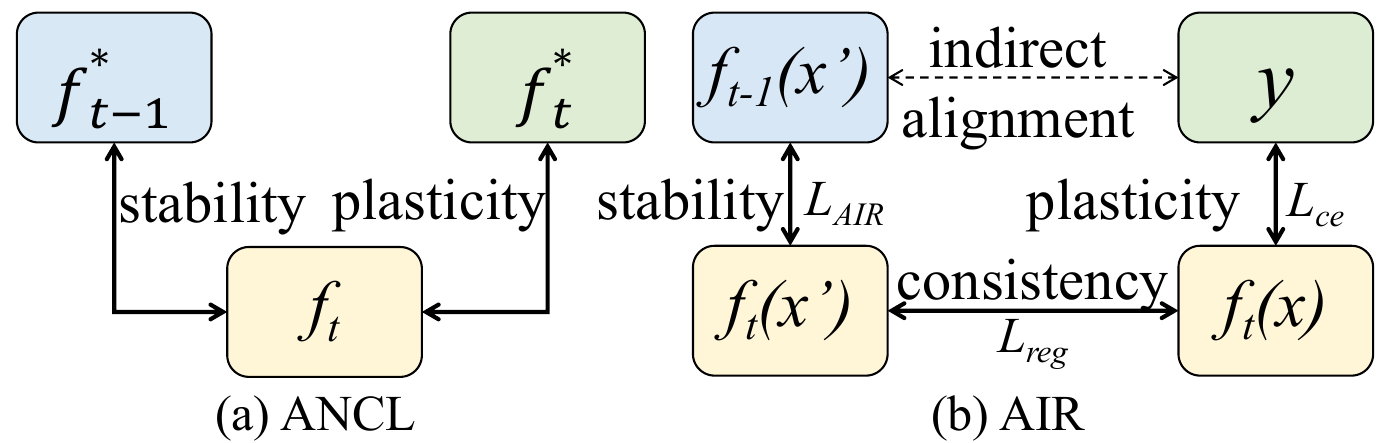}
\caption{Achievement of chain consistency of our AIR in end-to-end paradigm. Models with $*$ superscripts (such as $f^{*}_{t}$) are additionally trained independently of the main pipeline.}
\label{fig:reg}
\end{figure}


A common dilemma in continual learning is the trade-off between stability and plasticity. For the 'domain-incremental-like' continuous defense, a natural shortcut to optimizing the trade-off is to assign all attacks under a category to the same cluster. This approach elegantly optimizes the trade-off and achieves consistent optimization of attack sequences. In our AIR, augmentation data is considered as replay data to query the previous model and can also be considered as a neighborhood sample of new data. We propose aligning the outputs of the two as follows:

\begin{equation}
\begin{aligned}
\mathcal{L}_{reg} =\frac{1}{2}(KL(f_{w_t}^1(x_t)||&f_{w_t}^1(x_t^{'})) \\ & + KL(f_{w_t}^2(x_t)||f_{w_t}^2(x_t^{'}))),
\end{aligned}
\label{eq2}
\end{equation}
where $x_t^{'}$ is the isotropic augmentation replay data. The alignment process is implemented in the R-Drop~\cite{wu2021r} way, where $f_{w_t}^1$ and $f_{w_t}^2$ represent that the inputs are fed into the model with random Dropout~\cite{srivastavan2014dropout} twice, respectively.
On the one hand, a model with R-Drop will learn consistent outputs from different local features without overfitting on specific features. On the other hand, this constraint achieves an indirect chain alignment. Compared to the regularization from middle to the both sides in ANCL~\cite{kim2023achieving}, it resembles an alignment from both sides to the middle. Besides, AIR does not require additional auxiliary networks, and the alignment can be implemented in an end-to-end process (see in Figure~\ref{fig:reg}). The R-Drop is also applied to the AT for new attack with a minor probability.

\subsection{Final Model} \label{fin}
Integrally, the overall loss of AIR can be formalized as:
\begin{equation}
\begin{aligned}
\mathcal{L}_{AIR} = \mathcal{L}_{AT} + \lambda_{SD} \cdot (\mathcal{L}_{IR} + \mathcal{L}_{AR}) + \lambda_{Reg} \cdot \mathcal{L}_{Reg},
\end{aligned}
\label{eq2}
\end{equation}
where $\lambda_{SD}$ and $\lambda_{Reg}$ represent the hyper-parameters for self-distillation and regularization respectively. The AIR model will adapt to new attack while aligning with the previous model on the augmentation neighbor data, and the overall structure of AIR can be seen in Figure~\ref{fig:air}.

\section{Experiments}
\subsection{Experimental Setup}
\textbf{Datasets and Backbones.} Three commonly used datasets for adversarial attack \& defense are explored:

$\bullet$ MNIST: Following the settings in~\cite{zhang2019theoretically,carlini2017towards}, we employ the smallCNN architecture, consisting of four convolutional layers and three fully-connected layers. Typically, we set the perturbation parameter $\epsilon = 0.3$, perturbation step size $\eta_1 = 0.01$, number of iterations $K=40$, learning rate $\eta_2 = 0.01$, and batch size $m = 128$.

$\bullet$ CIFAR10 / CIFAR100: Following the settings in~\cite{zhang2019theoretically,madry2017towards}, we employ the WRN-34-10 and WRN-34-20 architecture~\cite{zagoruyko2016wide} for CIFAR10 and CIFAR100 respectively. We set the perturbation parameter $\epsilon = 8/255$, perturbation step size $\eta_1 = 2/255$, number of iterations $K = 10$, learning rate $\eta_2 = 0.1$, and batch size $m = 128$.

\noindent\textbf{Evaluation Protocol.}
We set attack sequences with lengths of 2 and 3, following the setting of existing continual learning works~\cite{rannen2017encoder,jung2016less,li2017learning}, including 'from hard to easy attack' and 'from easy to hard attack' strategies. The classic FGSM attack~\cite{goodfellow2014explaining} is selected as the easy attack, while the PGD attack~\cite{madry2017towards} is chosen as the hard attack. Considering that all attacks originate from benign samples, these benign samples are also considered as a base task. The adversarial training corresponding to these above two types of attacks serves as the adaptation processes to specific attacks. We do not choose the advanced AutoAttack~\cite{croce2020reliable}, as it is commonly used for evaluation rather than defensive training, and there is no specifically designed adaptation method for AutoAttack  to our best knowledge yet. Additionally, we will explore attack sequences with different budgets of PGD.

\subsection{Experimental Results}
\textbf{Attack Sequences of Different Lengths.} The continual defense results of the attack sequences with a length of 2 are shown in Table~\ref{tab1}, while that of the attack sequences with a length of 3 are shown in Table~\ref{tab2} (for MNIST), Table~\ref{tab3} (for CIFAR10) and Table~\ref{tab4} (for CIFAR100). Regardless of the length or arrangement of the attack sequence, the vanilla model consistently suffers from catastrophic forgetting. The internal maximization pattern and the 'ridges' reached by different attacks may vary, leading to a distribution shift in model parameters. Additionally, the 'from difficult to easy' attack sequence appears to exhibit more catastrophic forgetting compared to the 'from easy to difficult' sequence. This discrepancy may be attributed  to the adaptation process to simple attacks being more trivial, making the model more prone to overfitting and trapping in local optima. According to~\cite{ross2018improving}, the essence of adversarial training is smoothness regularization, and our AIR may have added stronger smoothness constraints compared on the basic traditional one-shot adversarial training to prevent output shifts in the case of continual input shift. Another issue is that the performance of our AIR may fluctuate. One can select a better model by monitoring the training online. A more detailed analysis of the results can be found in Section~\textcolor{red}{2} of the supplementary materials.

\begin{table}[]
\footnotesize
\centering
\setlength{\tabcolsep}{1mm}{
\begin{tabular}{c|ccc|ccc}
\toprule
\multicolumn{7}{c}{MNIST: None \& FGSM \& PGD}                                                                 \\ \midrule
\multirow{2}{*}{Tasks} & \multicolumn{3}{c|}{None to FGSM to PGD} & \multicolumn{3}{c}{PGD to FGSM to None} \\
                       & Task1          & Task2          & Task3          & Task1          & Task2          & Task3       \\ \midrule
Vanilla AT \cite{madry2017towards}               & 95.97          & 97.55          & 96.90          & 2.66           & 71.49          & 98.73       \\ \midrule
EWC \cite{kirkpatrick2017overcoming}                    & 98.97          & 96.47          & 92.91          & 89.67          & 95.43          & 99.11       \\
Feat. Extra \cite{li2017learning}    & 11.71          & 11.35          & 11.38          & 90.92          & 95.99          & 99.18       \\
LFL \cite{jung2016less}                    & 99.36          & 94.80          & 88.15          & 10.06          & 89.51          & 98.97       \\
\textbf{AIR (ours)}           & \textbf{99.39} & \textbf{97.21} & \textbf{94.54 }& \textbf{91.55 }& \textbf{97.34} & \textbf{99.33}       \\  \midrule
\cellcolor{lightgray}Joint Training \cite{li2017learning}            & \cellcolor{lightgray} 99.11          & \cellcolor{lightgray} 97.07          & \cellcolor{lightgray}94.76 & \cellcolor{lightgray}\textbf{94.76} & \cellcolor{lightgray} 97.07          & \cellcolor{lightgray}99.11       \\\bottomrule
\end{tabular}}
\caption{Results among None, FGSM, and PGD on MNIST.}
\label{tab2}
\end{table}

\begin{table}[]
\footnotesize
\centering
\setlength{\tabcolsep}{0.5mm}{
\begin{tabular}{c|ccc|ccc}
\toprule
\multicolumn{7}{c}{CIFAR10: None \& FGSM \& PGD}                                                               \\ \midrule
\multirow{2}{*}{Tasks} & \multicolumn{3}{c|}{None to FGSM to PGD} & \multicolumn{3}{c}{PGD to FGSM to None} \\
                       & Task1          & Task2          & Task3          & Task1          & Task2          & Task3       \\ \midrule
Vanilla AT \cite{madry2017towards}               & 70.93          & 40.70          & 44.04          & 21.87          & 36.80          & 84.59       \\ \midrule
EWC \cite{kirkpatrick2017overcoming}                    & 68.31          & 45.09          & 36.57          & 27.32          & 51.66          & 75.22       \\
Feat. Extraction \cite{li2017learning}     & 37.95          & 53.28          & 40.62          & \textbf{44.42} & \textbf{53.63} & 73.77       \\
LFL \cite{jung2016less}                    & 74.21          & 52.42          & 42.89          & 22.12          & 46.27          & 76.37          \\
\textbf{AIR (ours)}                    & \textbf{75.75} & \textbf{53.51} & \textbf{43.12} & 42.35          & 52.44          & \textbf{76.66}           \\ \midrule
\cellcolor{lightgray}Joint Training \cite{li2017learning}           & \cellcolor{lightgray}70.62          & \cellcolor{lightgray}51.35          & \cellcolor{lightgray}44.36          & \cellcolor{lightgray}44.36          & \cellcolor{lightgray}51.35          & \cellcolor{lightgray}70.62       \\ \bottomrule
\end{tabular}}
\caption{Results among None, FGSM, and PGD on CIFAR10.}
\label{tab3}
\end{table}

\begin{table}[]
\footnotesize
\centering
\setlength{\tabcolsep}{0.5mm}{
\begin{tabular}{c|ccc|ccc}
\toprule
\multicolumn{7}{c}{CIFAR100: None \& FGSM \& PGD}                                                               \\ \midrule
\multirow{2}{*}{Tasks} & \multicolumn{3}{c|}{None to FGSM to PGD} & \multicolumn{3}{c}{PGD to FGSM to None} \\
                       & Task1      & Task2      & Task3      & Task1      & Task2      & Task3       \\ \midrule
Vanilla                & 42.01      & 22.30      & 17.80      & 19.50      & 22.62      & 47.04       \\ \midrule
EWC \cite{kirkpatrick2017overcoming}                    & \textbf{48.35}      & 22.54      & 16.59      & 19.88      & 24.48      & 44.83          \\
Feat. Extraction \cite{li2017learning}    & 2.17       & 1.67       & 4.63       & 21.01      & 24.84      & 41.88       \\
LFL \cite{jung2016less}                    & 29.97      & 10.39      & 8.96       & 19.94      & 24.36      & \textbf{45.41}          \\
\textbf{AIR (ours)}     & 47.08          & \textbf{27.34}          & \textbf{23.04}          & \textbf{23.12}          & \textbf{27.04}          & 44.16           \\ \midrule
\cellcolor{lightgray}Joint Training \cite{li2017learning}           & \cellcolor{lightgray}45.33      & \cellcolor{lightgray}30.23      & \cellcolor{lightgray}21.25      & \cellcolor{lightgray}21.25      & \cellcolor{lightgray}30.23      & \cellcolor{lightgray}45.33       \\ \bottomrule
\end{tabular}}
\caption{Results among None, FGSM, and PGD on CIFAR100.}
\label{tab4}
\vspace{-0.3cm}
\end{table}


\begin{figure*}
\centering
\vspace{-0.1cm}
\includegraphics[width = 14cm]{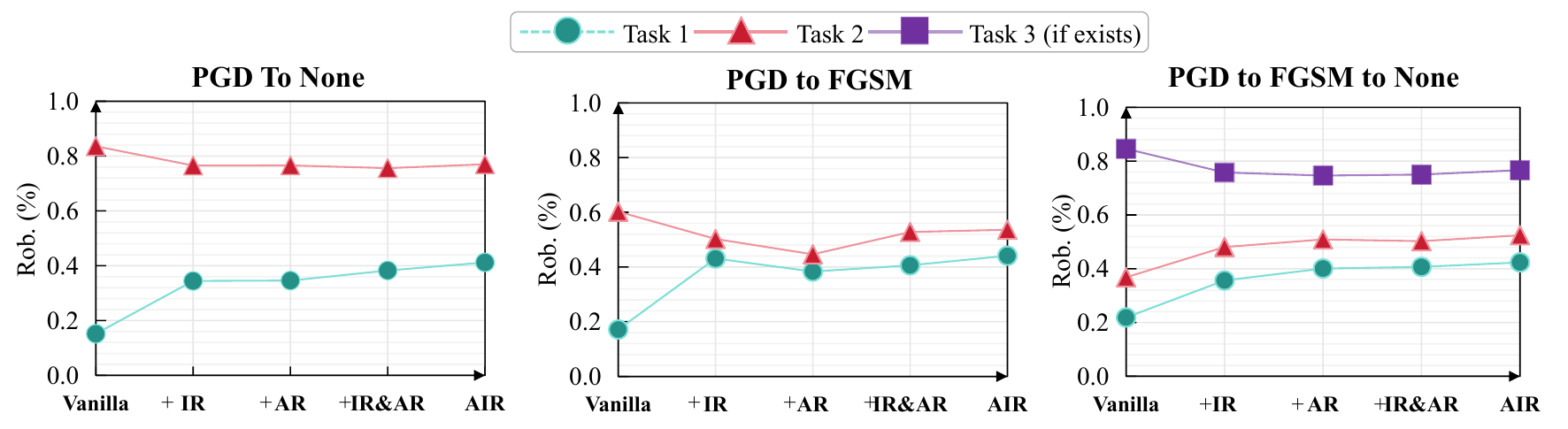}
\caption{Ablation analysis of the 'from hard to easy' attacks on CIFAR10. We reported its results after learning the whole attack sequence.}
\label{fig:ablation}
\vspace{-0.2cm}
\end{figure*}

\begin{figure}
\centering
\includegraphics[width = 6cm]{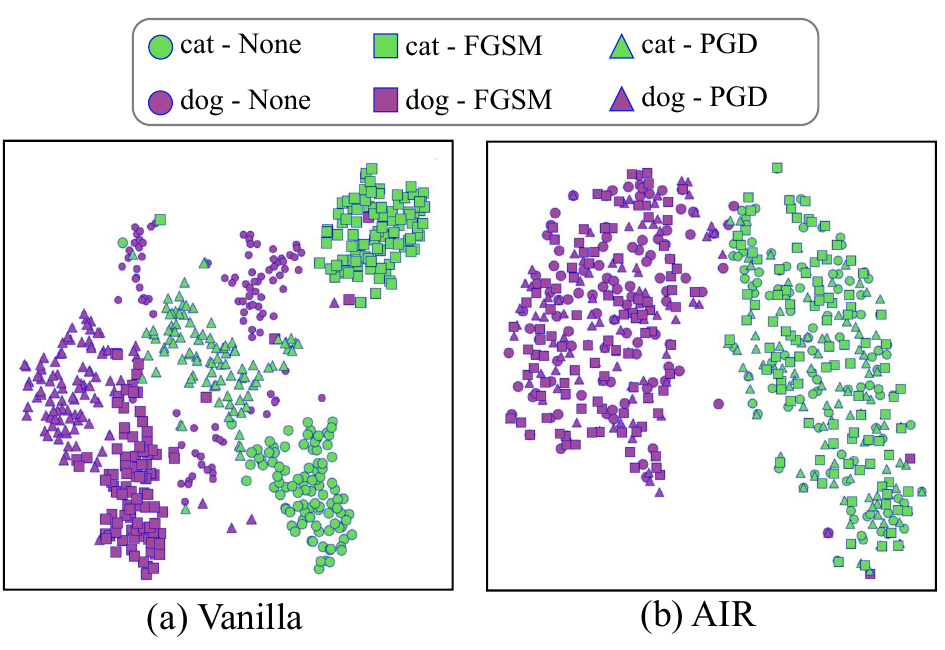}
\caption{T-SNE diagram of features encoded by vanilla AT and AIR on CIFAR10. The proposed AIR is able to encode all attacks in the sequence of the same category into one shared cluster.}
\label{fig:tsne}
\vspace{-0.5cm}
\end{figure}

\noindent\textbf{Transfer between different attack budgets.} Attacks with different budgets present different threats, which may seem counterintuitive. People may generally expect models adversarially trained with stronger internal maximization to be robust to weaker attacks. However, our findings reveal that the model exhibits different preferences for internal maximization with different budgets. This means that sequences composed of attacks with different intensities can lead to catastrophic forgetting. We briefly explore this issue on CIFAR10, setting attack budgets for strong and weak attacks to 8/255 and 80/255, respectively. Table~\ref{tab:intens} shows our experimental results and our AIR successfully alleviates the catastrophic forgetting caused by attack sequences with different attack budgets.

\begin{table}[]
\footnotesize
\centering
\setlength{\tabcolsep}{2.5mm}{
\begin{tabular}{c|cc|cc}
\toprule
\multicolumn{5}{c}{CIFAR10: Weak Attack \& Strong Attack}                                                    \\ \midrule
\multirow{2}{*}{Tasks} & \multicolumn{2}{c|}{Weak to Strong} & \multicolumn{2}{c}{Strong to Weak} \\
                      & Task1           & Task2          & Task1           & Task2          \\ \midrule
Vanilla               & 36.59           & 39.35          & 10.89           & 42.56          \\ 
\textbf{AIR (ours)}          & \textbf{45.70 }          & \textbf{37.20}          & \textbf{29.09}           & \textbf{42.87}       \\  
\cellcolor{lightgray}Joint Training \cite{li2017learning}       & \cellcolor{lightgray}37.03       & \cellcolor{lightgray}36.57 & \cellcolor{lightgray}36.57           & \cellcolor{lightgray}37.03          \\ \bottomrule
\end{tabular}}
\caption{Results between Weak \& Strong PGD on  CIFAR10.}
\label{tab:intens}
\vspace{-0.3cm}
\end{table}

\noindent\textbf{Ablation study.} 
We conducted ablation analysis of AIR on CIFAR10 with the 'hard to easy' attack sequence, which suffers from more serious forgetting and the comparisons are more significant. Figure~\ref{fig:ablation} shows the ablation study results. Essentially, each of our proposed modules contributes improvements for previous attacks (PGD attack in this case). The AR and IR modules individually enhance the performance against the previous attack, while the composite regularizer provides an overall increment for both previous and new attacks. Due to the trade-off involving 'plasticity-stability' and 'robustness-precision', our AIR, like other continual learning methods, also experiences performance degradation for new tasks. However, this sacrifice is necessary and cost-effective, as it results in effective improvement on previous attacks, aligning with the 'small pain but great gain' philosophy. Moreover, the overall performance of the model for both new and previous tasks steadily increases. In summary, the ablation analysis demonstrates the effectiveness of our designs in AIR.

\noindent\textbf{Discussion.} Our AIR sometimes can even outperforms Joint Training (JT). JT is typically considered as the empirical upper bound of continual learning, while our AIR can approach or even surpass JT in a memory-free manner. This may be attributed to the fact that the old attacks can be considered as the pre-training for new attacks. However, this superiority appears to be conditional: pre-training from easy to difficult attacks exhibits a better regularization effect on subsequent tasks, indicating that features from simpler tasks may be more generalizable. Utilizing previous task knowledge to enhance the learning of subsequent tasks may be one of the potentials of continual defense.

\subsection{Feature Distribution}
\textit{Our AIR tends to homogenize different attacks of the same category.} In one-shot vanilla training, the model often allocates different clusters for different attacks of the same class, as observed in Figure~\ref{fig:tsne}. This partially explains the forgetting mechanism in continual adversarial defense. While each class under the new attack is clustered, it does not share the same cluster as the previous attack. Consequently, attacks with the same label become isomerized. For instance, the features of FGSM and PGD attack of a certain label (e.g., 'dog') are assigned to different clusters, when ideally they should be in the same cluster. The adaptation learning of new clusters inevitably leads to the forgetting of old clusters. Intuitively, our AIR aligns the feature distribution of different attacks belonging to the same label into one cluster. This may benefit from the implicit alignment in the IR module and the chain regularizer, which aligns the output of new and previous models on the neighbor of the new data. This indirect alignment further harmonizes the feature distribution of new and previous models for different attacks of the same category, which provides an explanation beyond parameters regularization for AIR.

\subsection{The Twinship Between Two Trade-offs}

A common dilemma in adversarial defense is the trade-off between accuracy and robustness. Similarly, the 'plasticity-stability' dilemma in continual adversarial defense offers a novel perspective on the 'accuracy-robustness' trade-off. As the model transfers from benign data to adversarial data, the accuracy of benign tasks inevitably declines, reflecting the 'accuracy-robustness' dilemma. This reveals that the two dilemmas share similar insights: an excessive attention on the min-max process causes the model to forget relatively easy benign samples. Through the indirect alignment of the output preferences of old and new models, the pseudo-replay framework in our AIR can alleviate the forgetting problem of benign samples without accessing the original data. This alignment can be interpreted as optimizing the trade-off between accuracy and robustness. Such unified perspective reveals that ideas to alleviate the 'accuracy-robustness' dilemma may also be effective in mitigating the forgetting problem in continual adversarial defense. In defense communities, a common way to address the 'accuracy-robustness' trade-off is TRADES~\cite{zhang2019theoretically}, which aligns the output of clean and adversarial samples. However, querying previous data in continual defense is not feasible, making explicit alignment process challenging. Actually, our analysis indicates that AIR aligns the output preferences of new and old attacks in an indirect chain-like manner, which is similar to an implicit form of TRADES.

\section{Conclusion and Outlook}
In this paper, we first explore the challenge of achieving continual adversarial robustness under attack sequences, and verify that adaptation to new attacks can lead to catastrophic forgetting of previous attacks. Subsequently, we propose AIR as a memory-free continual adversarial defense baseline model. AIR aligns the outputs of new and old models in the neighborhood distribution of current samples and learns richer mixed semantic combinations to enhance adaptability to unknown semantics. An intuitive but efficient regularizer optimizes the generalization of multi-trade-offs in a chain-like manner.

One limitation of AIR is that it overlooks the regularization effect of previous knowledge for new tasks, where the previous tasks may act as pre-training. Additionally, we observed that in the 'from easy to difficult' attack sequence, AIR sometimes performs better in new tasks. There remains significant research space for continual adversarial defense.

\section{Acknowledgments}
This work was supported by the National Natural Science Foundation of China under Grant 62071142, the Shenzhen Science and Technology Program under Grant ZDSYS20210623091809029, and by the Guangdong Provincial Key Laboratory of Novel Security Intelligence Technologies under Grant 2022B1212010005.

{
    \small
    \bibliographystyle{ieeenat_fullname}
    \bibliography{main}
}
\end{document}